# Evolutionary Feature-wise Thresholding for Binary Representation of NLP Embeddings

Soumen Sinha[1], Shahryar Rahnamayan[2], and Azam Asilian Bidgoli[3]

[1]Department of EEMCS, TU Delft, Mekelweg 2628 CD, Delft Netherlands
[2]Department of Engineering, Brock University, St. Catharines, ON L2S 3A1, Canada.
[3]Faculty of Science, Wilfrid Laurier University, Waterloo, ON N2L 3C5, Canada
[*]Corresponding author: `S.Sinha-6@student.tudelft.nl`

**Abstract**

Efficient text embedding is crucial for large-scale natural language processing (NLP) applications, where storage and computational efficiency are key concerns. In this paper, we explore how using binary representations (barcodes) instead of real-valued features can be used for NLP embeddings derived from machine learning models such as BERT. Thresholding is a common method for converting continuous embeddings into binary representations, often using a fixed threshold across all features. We propose a Coordinate Search-based optimization framework that instead identifies the optimal threshold for each feature, demonstrating that feature-specific thresholds lead to improved performance in binary encoding. This ensures that the binary representations are both accurate and efficient, enhancing performance across various features. Our optimal barcode representations have shown promising results in various NLP applications, demonstrating their potential to transform text representation. We conducted extensive experiments and statistical tests on different NLP tasks and datasets to evaluate our approach and compare it to other thresholding methods. Binary embeddings generated using using optimal thresholds found by our method outperform traditional binarization methods in accuracy. This technique for generating binary representations is versatile and can be applied to any features, not just limited to NLP embeddings, making it useful for a wide range of domains in machine learning applications.

# 1. Introduction

Natural Language Processing (NLP) has become an essential component of various applications, including machine translation, sentiment analysis, question-answering systems, and information retrieval. The ability of machines to understand and process human language efficiently has led to significant advancements in AI-driven technologies. At the core of many NLP models lies the concept of embeddings, which are vectorized representations of words, sentences, or documents in a continuous space. These embeddings enable models to capture semantic relationships between words, improving their ability to understand context and meaning. However, as NLP models grow in complexity, managing high-dimensional embeddings poses computational and memory challenges, leading to research on techniques such as binarization of embeddings to optimize storage and processing efficiency. In recent years, binarization of embeddings [1, 2, 3, 4] has emerged as a significant approach to address the challenges of high-dimensional data representation in Natural Language Processing (NLP). The conversion of continuous embeddings into binary representations, also known as barcode representation, has demonstrated notable advantages in terms of memory savings and computational efficiency.

Binarization of embeddings involves transforming real-valued features into binary codes, which reduces storage requirements and accelerates operations in various machine learning models. This method is particularly beneficial for resource-intensive, large-scale text processing tasks. Notable methods in this domain include Radon barcodes [5], Min-max Radon barcodes [6], and auto-encoded Radon barcodes [7, 8]. These approaches leverage binary formats to optimize data retrieval and processing. The introduction of transformer-based models like BERT (Bidirectional Encoder Representations from Transformers) by Devlin et al. [9] has revolutionized NLP tasks by learning rich contextual embeddings of text. However, managing the high-dimensional continuous embeddings from models like BERT can be computationally and memory-intensive. This challenge has led researchers to explore various binarization techniques to mitigate these issues. The impact of binarization is significant, as it enables the deployment of NLP models on devices with limited computational resources by reducing memory usage and enhancing processing speed. For instance, Tissier et al. [10] proposed a near-lossless binarization method that compresses word embeddings to a fraction of their original size while preserving semantic information. Similarly, Navali et al. [11] introduced a technique that converts continuous embeddings into binary representations, achieving substantial reductions in file size with minimal impact on performance. Addressing the necessity of binarization, the growing complexity and size of NLP models necessitate efficient storage and faster computation, especially for applications on mobile devices and real-time systems. Binarization techniques offer a viable solution to these challenges, ensuring that advanced NLP models remain accessible and practical across various platforms.

Moreover, binarization plays a vital role in the democratization of NLP by enabling the deployment of language models on low-power devices such as smartphones, embedded systems, and IoT platforms [12]. As foundation models continue to grow in size and complexity, conventional float-based representations become prohibitively expensive for real-time inference. Binary neural networks (BNNs) and discrete embeddings offer an attractive trade-off between model fidelity and latency [13]. Recent studies have also explored the synergy between binarization and approximate nearest neighbor (ANN) search in large-scale retrieval systems [14], demonstrating that binary representations can significantly speed up indexing while maintaining competitive retrieval accuracy. Furthermore, research in binarized transformers [15] suggests that even end-to-end transformer architectures can benefit from binary quantization without major losses in accuracy, emphasizing that binarization is not merely a storage trick, but a viable architectural choice. These trends point toward a broader paradigm shift where binarization is treated not just as a downstream compression method, but as an integral part of model design in resource-aware machine learning.

Researchers exploring binary embeddings have sought innovative solutions to address the need for fast retrieval and low memory complexity in data representation. One such solution involves the use of barcodes as an alternative to traditional data features. These barcode-based methods essentially provide binary representations, aligning with the core idea of binary embeddings. Prominent examples of such methods include Radon barcodes [16], Minmax radon barcodes [17], SVM-based approaches [18], and auto-encoded Radon barcodes. The significance of these barcode-based methods lies in their ability to represent data features in a binary format, which not only reduces memory requirements but also facilitates faster retrieval. Knipper et al. [19] and Mostard et al. [20] proposed various methods to acheive binary embeddings. Mostard et al. introduce a Siamese autoencoder for semantic hashing of word embeddings into binary form with minimal information loss and Knipper et al. introduced a genetic algorithm that learns relationships between words from existing word analogy datasets. Another intresting work was proposed by Hamidreza et al. [21] which involved Content-Based histopathology image retrieval with the help of QR code representation.

The translation of complex features into binary codes has been a challenging task in data representation. Conventionally, one of the widely adopted methods for this purpose is thresholding, a process that involves the conversion of continuous data into binary form. Various thresholding techniques have been proposed to achieve optimal binary representations, including simple thresholding, MinMax thresholding [22], and Otsu thresholding [23]. Although these methods are widely used, determining an optimal threshold value remains a challenging task. Threshold values are critical in thresholding methods, as they define how continuous data is mapped to binary representations. Determining the optimal threshold value can be formulated as an optimization problem to generate more accurate binary embeddings, particularly for tasks such as text classification. A fixed threshold across all features is common but often suboptimal, potentially leading to information loss or noise. Optimizing thresholds per feature enhances representation quality, improving accuracy in tasks like classification and retrieval.

The Coordinate Search (CS) algorithm [24] is a straightforward yet effective optimization technique that iteratively adjusts individual parameters to locate an optimal solu-tion. Unlike gradient-based methods, CS does not require derivative information, making it well-suited for non-differentiable or complex objective functions. It operates by sequentially varying one coordinate at a time while keeping others fixed, progressively refining the solution.

In this context, the CS algorithm efficiently explores the threshold search space, identifying the optimal value for each feature that enhances binary embedding quality. By systematically optimizing the threshold, CS improves text processing accuracy, making it a practical choice for refining embeddings in classification and retrieval tasks. This method has been successfully applied in various optimization problems, including iterative shrinkage/thresholding algorithms for sparse solutions [11], and training artificial neural networks [25]. These methods tackle optimization problems by solving a series of simpler sub-problems. The apparent simplicity and satisfactory performance of the coordinate search approach in various scenarios likely contribute to its enduring popularity among practitioners. Notable contributions in this area include an accelerated proximal coordinate gradient method by Lin et al. [26], and coordinate descent algorithms for nonconvex penalized regression by Breheny and Huang [27]. The application of evolutionary computation techniques in optimization has also been explored extensively. Bidgoli and Rahnamayan introduced Memetic Differential Evolution with Coordinate Search [28], while Ehsan et al. presented a coordinate search algorithm for training artificial neural networks [29]. Additionally, Chang et al. [30] proposed a coordinate descent approach tailored for large-scale l2-loss linear support vector machines, aiding in the optimization challenges of support vector machines in machine learning.

## 2. Background Review

Within the domain of data representation, translating complex features into binary codes has been a challenging task. Conventionally, one of the widely adopted methods for this purpose is thresholding, a process that involves the conversion of continuous data into binary form. In following subsections, the well-known thresholding methods that were used or compared with the proposed method are reviewed.

### 2.1 Simple Thresholding

Simple thresholding [31] involves selecting a fixed threshold value ($T$) and assigning binary values based on whether feature values exceed or are below the threshold. The formula for simple thresholding is as follows:

$$B_i = \begin{cases} 0, & \text{if } X_i < T, \\ 1, & \text{if } X_i \geq T. \end{cases} \quad (1)$$

Where $B_i$ is the binary value of the $i$-th feature. $X_i$ is the $i$-th feature value and $T$ is the threshold value. Threshold value $T$ is used on all the features in the training data.

## 2.2 MinMax Thresholding

We also applied MinMax thresholding, a technique introduced by Jain and Zongker [22], for feature selection and data transformation. It is an approach that considers the relative changes between consecutive feature values to determine the binary representation. If a feature value in the vector exceeds its preceding feature, it is assigned a binary **1**; otherwise, it receives a binary **0**. The MinMax thresholding formula is expressed as:

$$B_i = \begin{cases} 1, & \text{if } X_i > X_{i-1}, \\ 0, & \text{if } X_i \leq X_{i-1}. \end{cases} \quad (2)$$

Where $B_i$ is the binary value of the $i$-th feature. $X_i$ is the $i$-th feature value and $X_{i-1}$ is the value of the preceding feature.

## 2.3 Otsu Thresholding

Otsu's method [32] is a well-known thresholding technique used to separate data into two classes by finding an optimal threshold ($T$) that minimizes the intra-class variance while maximizing the inter-class variance. The formula for Otsu's thresholding is as follows:

$$T_{\text{Otsu}} = \arg\max_{T} \left\{ \frac{\sigma_B^2}{\sigma_W^2} \right\}. \quad (3)$$

Where $T_{\text{Otsu}}$ is the optimal threshold. $\sigma_B^2$ is the inter-class variance and $\sigma_W^2$ is the intra-class variance. After obtaining $T_{\text{Otsu}}$, it is applied to all the features in the training data. It is important to note that otsu thresholding is a optimization based method.

## 2.4 Hybrid Thresholding

Hybrid thresholding [33] combines MinMax scaling with mean ($\mu$) and median ($\tilde{x}$) values to convert continuous embeddings into binary representations. The threshold ($T$) is computed as the average of $\mu$ and $\tilde{x}$. Elements exceeding $T$ become **1**, while those below become **0**. The process can be summarized as:

$$T = \frac{\mu + \tilde{x}}{2}. \quad (4)$$

Binary Embedding ($B_i$) for element $i$:

$$B_i = \begin{cases} 1, & \text{if } X_i > T, \\ 0, & \text{if } X_i \leq T. \end{cases} \quad (5)$$

Hybrid thresholding offers a flexible approach for binary conversion of embeddings. After obtaining $T$, it is applied to all the features in the training data behaving as a global threshold.

## 2.5 BERT: Bidirectional Encoder Representations from Transformers

BERT (Bidirectional Encoder Representations from Transformers) is a deep learning model introduced by Devlin et al. [34] that has significantly advanced the state-of-the-art in a wide range of natural language processing (NLP) tasks. Unlike traditional models that read text sequences from left to right or right to left, BERT leverages the Transformer architecture to learn context from both directions simultaneously. This bidirectional pre-training allows BERT to capture deep semantic and syntactic relationships between words.

BERT is pretrained on large text corpora using two unsupervised tasks: Masked Language Modeling (MLM), where random tokens in a sentence are masked and the model learns to predict them, and Next Sentence Prediction (NSP), where the model learns to understand the relationship between sentence pairs. Once pretrained, BERT can be fine-tuned with minimal architecture changes for various downstream tasks such as sentiment analysis, question answering, and named entity recognition, achieving state-of-the-art performance with minimal task-specific modifications.

## 2.6 Coordinate Search Algorithm

Coordinate Descent (CD) algorithms are a class of decomposition-based optimization techniques that solve complex problems by iteratively optimizing a single coordinate or a block of coordinates while keeping others fixed [35, 36]. When gradient information is unavailable or expensive to compute, CD algorithms can substitute full derivatives with one-dimensional searches along individual coordinate directions [37]. In numerical linear algebra, gradients are typically required; however, in the context of Evolutionary Computation, the derivative-free counterpart is known as Coordinate Search (CS). CS leverages function evaluations along coordinate directions, sequentially updating variables to improve the solution.

Despite its conceptual simplicity, CS is remarkably effective in practice particularly for computationally expensive tasks like feature selection. Its flexibility allows for various configurations depending on parameters such as coordinate order, number of coordinates updated per iteration, sampling strategy, and initialization [38]. To accelerate convergence in high-dimensional settings, block-wise CS variants can update multiple variables simultaneously, which significantly reduces the number of required evaluations while maintaining solution quality.

# 3. Proposed Method

In this study, we introduce a novel approach to convert continuous BERT embeddings into binary embeddings by determining optimal thresholds for each feature. This method retains essential information from the word embeddings while significantly reducing computational and memory requirements. Effective binary representation of continuous embeddings relies on selecting appropriate threshold values for each feature. Traditional methods often apply a fixed threshold across all features, which can lead to suboptimal performance. To address this, we propose a CS-based optimization framework that determines the optimal threshold for each feature individually, improving the quality of the binary encoding.

Our approach formulates threshold selection as an optimization problem, where the objective is to maximize a fitness function that evaluates the quality of the binary representation (e.g., text classification accuracy). The CS algorithm iteratively explores threshold values for each feature, refining them to achieve an optimal solution. The optimization process involves evaluating the fitness of candidate solutions and updating thresholds accordingly. In the following sections, we describe the details of the search strategy. Our approach involves three main steps: data preprocessing, threshold optimization using the CS algorithm, and binary embedding generation.

Initially, we preprocess the text data by tokenizing it and then obtain the embeddings using the BERT model. Unlike traditional models, BERT generates contextually rich embeddings by considering the bidirectional context of words in a sentence. The core of BERT's architecture is based on the transformer model, which relies on attention mechanisms to capture dependencies between input and output.

The next critical step in processing word embeddings for binary classification tasks is threshold optimization. This process involves determining an effective threshold vector ($S^*$) that maximizes classification performance. While continuous embeddings like those generated by BERT offer rich semantic information, they come at the cost of high memory and computational demands. Binarization serves as a practical solution, compressing real-valued embeddings into binary codes, which reduces storage overhead and speeds up inference through efficient bitwise operations. However, naively applying a global or uniform threshold can lead to information loss or distortion, degrading the performance of downstream classifiers. To preserve semantic fidelity while gaining the benefits of binarization, it is essential to carefully calibrate the thresholds used for each feature dimension. This calibration ensures that the resulting binary embeddings approximate the structure and discriminative power of their continuous counterparts. The threshold optimization process involves determining an effective threshold vector ($S^*$) that maximizes classification performance. Metrics such as accuracy, F1-score, or AUC can guide this optimization. In this context, the CS algorithm plays a pivotal role. As a derivative-free optimization method, CS sequentially refines one threshold coordinate at a time, exploring the search space between lower and upper bounds for each dimension. The goal is to minimize the discrepancy between the continuous and binary representations, ensuring that the binary embeddings maintain task-specific utility while offering computational efficiency. The proposed algorithm is outlined in Algorithm 1.

## 3.1 Threshold Optimization

The CS algorithm is an iterative method used to find the optimal threshold for each feature to convert continuous embeddings to binary embeddings. Beginning with an initial threshold vector $S^*$, the algorithm also utilizes defined lower ($L$) and upper ($U$) bounds for each dimension. It executes for a predefined maximum number of iterations ($maxiter$).

The first specific operation of the CS algorithm is to find the region of interest (ROI). For each dimension, represented as $i$, the search space is defined as $[L_i, U_i]$. To select each dimension, a permutation vector is utilized. During each iteration to find the region of interest, the search space for the $i$-th dimension is divided into two equal parts, creating two regions: $[L_i, \frac{L_i+U_i}{2})$ and $[\frac{L_i+U_i}{2}, U_i]$. To identify a representative for each region, the algorithm suggests using the center of each region as the decision variable. This choice is motivated by the concept of center-based sampling strategy [39]. Studies have shown that the likelihood of a center point being close to a random solution is much higher compared to random points, especially for large-scale problems. Consequently, the center point serves as a suitable representative for each region. In mathematical terms, the representatives for each interval (center points) are computed as $L_i + \frac{U_i - L_i}{4}$ and $U_i - \frac{U_i - L_i}{4}$.

These computed center points are then assigned to the $i$-th dimension of two candidate thresholds. Subsequently, the F1-scores of classification are calculated for both thresholds. The candidate threshold with the superior F1-score value is chosen to determine the interval of interest for the $i$-th dimension.

After identifying the interval of interest, the next operation is to shrink the search space. To achieve this, the upper and lower bounds of each dimension are adjusted to match the upper and lower bounds of the interval of interest. These two steps, finding the interval of interest and shrinking the search space, are performed for each dimension independently. In each iteration, the algorithm selects a random dimension $i$ and computes two candidate threshold values $X_i^{(1)}$ and $Y_i^{(2)}$ for this dimension. These values are computed as follows:

$$X_i^{(1)} = L[i] + \frac{U[i] - L[i]}{4}, \tag{6}$$

$$Y_i^{(2)} = U[i] - \frac{U[i] - L[i]}{4}. \tag{7}$$

These candidate thresholds are potential new values for converting each of the continuous embeddings into a binary format. The algorithm then evaluates the performance of each threshold by calculating the F1-score. Let $F1_X$ and $F1_Y$ denote the F1-scores obtained by applying $X_i^{(1)}$ and $Y_i^{(2)}$, respectively. After iterating through all dimensions for *maxiter* iterations, the algorithm yields an optimized threshold vector $S^*$. This optimized vector is then used to transform the continuous embeddings into binary embeddings, suitable for downstream tasks such as sentiment analysis using logistic regression classifiers. The optimization process ensures that the binary embeddings retain the essential characteristics of the original continuous embeddings, thereby enhancing the performance of classification models. In our case, we considered the upper bound to be 1 and the lower bound to be -1, as the BERT embeddings of our dataset lie within this range.

**Algorithm 1** CS Optimization for Thresholding Embeddings
---
**Require:** embeddings: Input word embeddings, maxiter: Maximum number of iterations
**Ensure:** $S^*$: Optimized binary threshold vector
    $num\_samples, D \leftarrow \text{shape}(embeddings)$     ▷ Number of samples and embedding dimension
    $L \leftarrow \text{ones}(D) \times -1$     ▷ Initialize lower bound
    $U \leftarrow \text{ones}(D)$     ▷ Initialize upper bound
    $S^* \leftarrow \text{zeros}(D)$     ▷ Initialize best solution
    $MaxNFE \leftarrow num\_samples \times maxiter \times 2$
    $R_{\max} \leftarrow MaxNFE / (2 \times D \times maxiter)$
    **for** $R = 1$ **to** $R_{\max}$ **do**
        $X \leftarrow 0.5 \times (L + U)$     ▷ Initialize X
        $Y \leftarrow X$     ▷ Initialize Y
        $Perm \leftarrow \text{rand\_perm}(D)$     ▷ Random permutation of dimensions
        **for** iter $= 1$ **to** maxiter **do**
            **for** ind $= 1$ **to** $D$ **do**
                $i \leftarrow Perm[\text{ind}]$
                $C \leftarrow 0.5 \times (L[i] + U[i])$
                $q \leftarrow 0.25 \times (U[i] - L[i])$
                $X[i] \leftarrow L[i] + q$
                $Y[i] \leftarrow U[i] - q$
                $binary\_X \leftarrow \text{ToBinary}(embeddings, X)$
                $binary\_Y \leftarrow \text{ToBinary}(embeddings, Y)$
                $F1_X \leftarrow \text{Evaluate}(binary\_X)$
                $F1_Y \leftarrow \text{Evaluate}(binary\_Y)$
                **if** $F1_X > F1_Y$ **then**
                    $S \leftarrow X$
                    $U[i] \leftarrow C$
                **else**
                    $S \leftarrow Y$
                    $L[i] \leftarrow C$
                **end if**
                $X \leftarrow S$
                $Y \leftarrow S$
            **end for**
        **end for**
        $binary\_S \leftarrow \text{ToBinary}(embeddings, S)$
        $F1_S \leftarrow \text{Evaluate}(binary\_S)$
        **if** $F1_S > F1_{S^*}$ **then**
            $S^* \leftarrow S$
            $F1_{S^*} \leftarrow F1_S$
        **end if**
    **end for**
    **return** $S^*$

To gain a better understanding of the CS algorithm and how it finds an optimal threshold for each feature, let's illustrate its operation with a simplified example involving an embedding including four variables: $x_1, x_2, x_3$, and $x_4$, without any permutations. In this scenario, the search space for all variables is set to $[-1, +1]$, and the algorithm is applied to solve a minimization problem.

In the initialization phase, two identical center-based candidate solutions, $X = (0, 0, 0, 0)$ and $Y = (0, 0, 0, 0)$, are generated. In the next step, the region of interest for $x_1$ should be calculated. Initially, the search space for this dimension is divided into two equal sub-regions: $[L_1 = -1, U_1 + \frac{L_1+U_1}{2} = 0]$ and $[\frac{L_1+U_1}{2} = 0, U_1 = +1]$. Figure 1 demonstrates the search space reduction in every interation. Consequently, the center points for these sub-regions are determined as $-0.5$ and $0.5$. Two candidate solutions, $X = (-0.5, 0, 0, 0)$ and $Y = (0.5, 0, 0, 0)$, are generated, where the values for the first dimension correspond to the center points ($-0.5$ and $0.5$). Assuming that the F1-scores using thresholds $X$ and $Y$ are $0.65$ and $0.52$, respectively, the winning candidate is $X$. Consequently, the new search space for the first dimension becomes $[-1, 0]$, while the search space for the other dimensions remains unchanged. This process is then repeated for the second dimension, $x_2$, and so on for the remaining dimensions.

It is important to note that in each iteration, the search space is shrunk by a factor of $\frac{1}{2D}$, where $D$ represents the number of dimensions. This exponential reduction of the search space in each iteration allows the CS algorithm to effectively converge. Additionally, the CS algorithm has the advantage of being parameter-free, eliminating the need for control parameter tuning. Figure 2 shows a representation of how search space is reduced with each iteration.

In the CS algorithm, the number of runs depends on the number of iterations and the maximum number of function evaluations. It is determined as $R_{\max} = \frac{\text{maxNFE}}{2 \times D \times \text{maxiter}}$, where $R_{\max}$ represents the maximum number of runs with different orders, maxNFE is the maximum number of function evaluations, and *maxiter* is the number of iterations.

Upon determining the optimal threshold vector $S^*$, we convert the real value embeddings into binary features.

Importantly, the CS algorithm leverages function evaluations instead of gradient information, making it highly suitable for non-differentiable, black-box optimization tasks such as binarizing embeddings. The exponential reduction in search space ensures that each dimension converges toward an optimal threshold, while also maintaining computational efficiency. In practice, the process terminates after a predefined number of iterations or when the improvement in the objective metric falls below a specified threshold. The final optimized threshold vector $S^*$ is a product of these localized decisions across all dimensions, aiming to preserve the discriminative power of the original continuous embeddings while benefiting from the reduced complexity of binary representations. Through this iterative, dimension-wise refinement, the CS algorithm achieves a robust and interpretable binarization scheme that is well-aligned with downstream classification objectives.

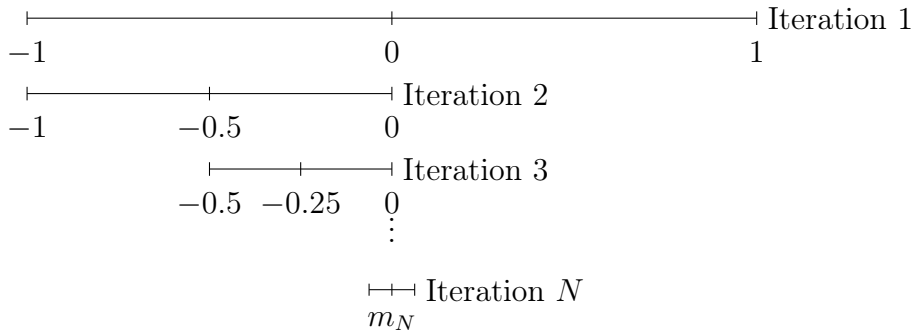

Figure 1: Visual representation of search space reduction. After $N$ iterations, $\mathcal{I}_N = [a_N, b_N]$ with length $b_N - a_N = 2^{1-N}$ and midpoint $m_N = \frac{a_N + b_N}{2}$.

The proposed method representation is shown in Figure 2

$$\text{Binary Embedding}[i] = \begin{cases} 1, & \text{if } \text{Real}_{\text{array}}[i] \geq S^*[i], \\ 0, & \text{otherwise.} \end{cases} \quad (8)$$

Here, $\text{Real}_{\text{array}}$ represents the array containing the real value embeddings and $S[i]$ represents the $i$-th element of the array of optimal thresholds.

These binary embeddings are then utilized to train a logistic regression classifier to evaluate the accuracy of the achieved binary representation for NLP tasks.

## 4. Experimental Analysis

### 4.1 Experimental Setup

We conducted experiments to evaluate the proposed method on various NLP datasets, including IMDb, GLUE SST-2, AG News, SNLI, and CoNLL. The experiments consisted of processes that were both global-based and feature-based. Specifically, the global-based processes involved applying a global threshold to all features, while in the feature-based processes thresholds were determined per feature. The two methods were further divided into optimizer-based and non-optimizer-based methods. Figure 3 explains the various techniques considered for evaluation.

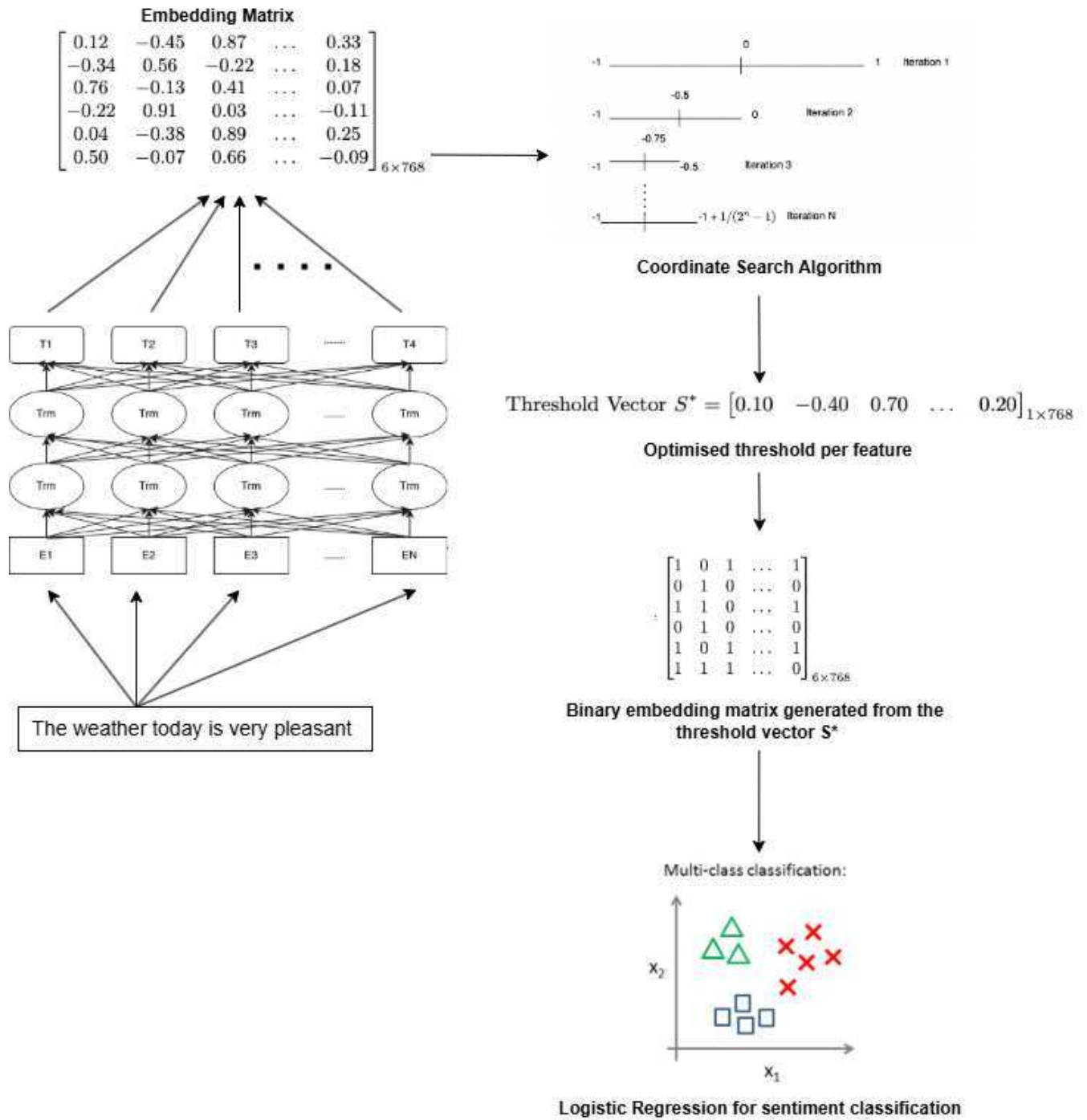

Figure 2: Flow of the proposed Method for text Classification. Input text is processed through BERT model to generate embeddings, which are then binarized using an optimized threshold from the Coordinate Search Algorithm.

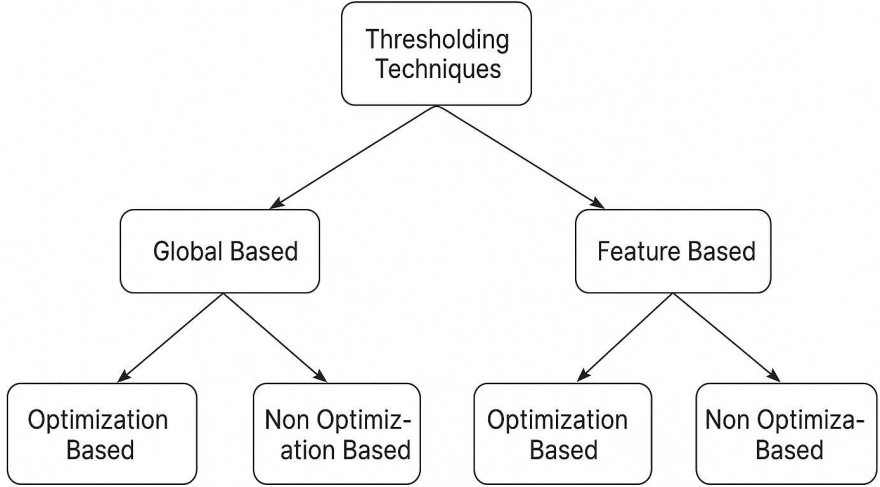

Figure 3: Variation in thresholding techniques considered during experimentation

Methods used during experimentation are as follows.

- **Global-based:** Simple, Hybrid, Otsu, CS-based Global

- **Feature/local based:** Min-Max, CS-based (proposed method)

All experiments were performed on the DGX-1 supercomputer platform. The DGX-1's kernel is a dual-core CPU server equipped with 20 processors and 8 Tesla V100 GPU cards, which collectively comprise 40,960 Nvidia CUDA cores. This setup provided the necessary computational power to efficiently process and analyze the large-scale NLP datasets.

## 4.2 Dataset Description

We conducted our experiments on 5 datasets namely IMDb, GLUE-sst2, AG-News, CoNLL-2003 and SNIL. Table. 1 outlines the specifications of various NLP datasets used in our experiments. The size of the extracted embedding for each sample is 768 for all datasets.

Table 1: Specification of various NLP datasets

| Dataset Name | Number of Samples | Number of Classes | Dimension of Embedding |
|---|---|---|---|
| IMDb | 50,000 | 2 | 768 |
| GLUE-sst2 | 70,000 | 2 | 768 |
| AG News | 120,000 | 4 | 768 |
| CoNLL-2003 | 34,266 | 4 | 768 |
| SNIL | 570,000 | 3 | 768 |

## 4.3 Experimental Results

Table 2: **Median classification accuracy (15 independent runs) ± standard deviation** of all thresholding strategies on five NLP benchmarks. "**CS-Feature**" is our proposed method which is per feature based, "CS-Global" applies coordinate search to a single global threshold, and the remaining rows are classical thresholding methods. Real-valued BERT embeddings (italic) act as an upper-bound reference to compare the binary embeddings with real valued embeddings. family. Best scores per dataset is in **bold**

| Method | IMDb (%) | GLUE SST-2 (%) | AG News (%) | CoNLL (%) | SNLI (%) |
|---|---|---|---|---|---|
| **Proposed Method** | **87.84 ± 0.45** | **82.70 ± 0.38** | **85.72 ± 0.42** | **77.74 ± 0.50** | **76.70 ± 0.47** |
| Real values embeddings (BERT) | 86.68 ± 0.40 | 84.77 ± 0.35 | 87.93 ± 0.38 | 82.93 ± 0.48 | 82.22 ± 0.43 |
| Optimization based CS Global | 81.64 ± 0.52 | 81.88 ± 0.47 | 83.67 ± 0.51 | 71.35 ± 0.55 | 54.35 ± 0.50 |
| Optimization based Simple | 80.12 ± 0.53 | 79.35 ± 0.50 | 82.34 ± 0.49 | 70.32 ± 0.52 | 70.12 ± 0.47 |
| Optimization based Otsu | 78.56 ± 0.60 | 77.23 ± 0.55 | 80.15 ± 0.58 | 68.45 ± 0.61 | 68.45 ± 0.59 |
| Optimization based Hybrid | 76.43 ± 0.62 | 75.68 ± 0.57 | 79.68 ± 0.61 | 66.88 ± 0.65 | 50.23 ± 0.63 |
| Simple | 66.31 ± 0.70 | 68.01 ± 0.66 | 66.34 ± 0.71 | 65.42 ± 0.68 | 71.17 ± 0.75 |
| MinMax [40] | 65.38 ± 0.75 | 66.36 ± 0.72 | 51.03 ± 0.78 | 54.24 ± 0.82 | 56.63 ± 0.77 |
| Hybrid [33] | 56.96 ± 0.85 | 66.88 ± 0.80 | 31.11 ± 0.89 | 28.19 ± 0.92 | 30.53 ± 0.86 |
| Otsu [41] | 46.57 ± 0.90 | 75.05 ± 0.86 | 75.05 ± 0.91 | 64.95 ± 0.88 | 64.95 ± 0.85 |

In Table 2 median accuracies for various methods are reported. The proposed method, which applies Coordinate Search in a feature-based manner using an optimizer, demonstrated superior performance across multiple datasets. Unlike **CS-Global**, which searches for a *single* threshold shared by all 768 embedding dimensions, our **proposed method** optimises *one threshold per dimension*. This extra granularity allows highly informative features to adopt relaxed cut-points while noisy ones receive stricter thresholds, yielding consistently higher accuracy and smaller variance. For *Optimised Simple*, *Optimised Otsu*, and *Optimised Hybrid* we start from the classical heuristic cut-point (0, Otsu's variance-ratio value, or the mean–median average) and then fine-tune that single scalar with a one-dimensional Coordinate-Search that maximises validation macro-$F_1$.

As shown in Table 2, the proposed method achieved the highest median accuracy on IMDb (**87.84%**), GLUE SST-2 (**82.70%**), AG News (**85.72%**), CoNLL (**77.74%**), and SNLI (**76.70%**).

In comparison, the Real values embeddings (BERT) method achieved median accuracies of 86.68% on IMDb, 84.77% on GLUE SST-2, 87.93% on AG News, 82.93% on CoNLL, and 82.22% on SNLI. Although the Real values embeddings (BERT) method outperformed proposed on GLUE SST-2, AG News, CoNLL, and SNLI datasets, the proposed method still showed competitive performance and achieved the highest accuracy on the IMDb dataset. In two scenarios (GLUE SST-2 and AG News), the difference in accuracies between the proposed and Real values embeddings (BERT) methods was minimal. This highlights that although real value embeddings are generally expected to perform better due to the richer information they contain, the binary embeddings produced by the proposed are closely approaching their performance.

This is particularly significant as the proposed method offers benefits in terms of memory savings and computational costs. Binary embeddings require less storage space and can be processed more efficiently, making the proposed method a valuable altern-ative in resource-constrained environments. The consistent performance of the proposed method across different datasets highlights its robustness and effectiveness. The method's ability to adapt to various types of data and maintain high accuracy demonstrates its potential for broader applications in NLP tasks. Furthermore, the optimization based CS-Global method, which applies the global threshold with an optimizer, showed competitive results but was outperformed by the feature-based proposed method in most cases. This underscores the importance of using feature-based optimization in enhancing the performance of Coordinate Search.

Table 3: Memory Usage and Computation Time for All Methods and Datasets

| Method | IMDb (MB/ms) | GLUE SST-2 (MB/ms) | AG News (MB/ms) | CoNLL-2003 (MB/ms) | SNLI (MB/ms) |
| --- | --- | --- | --- | --- | --- |
| Proposed Method | 4.98 / 120 | 7.02 / 130 | 9.01 / 140 | 3.51 / 110 | 10.03 / 150 |
| Real Embeddings (BERT) | 146.48 / 550 | 205.99 / 580 | 351.56 / 620 | 100.36 / 520 | 1669.92 / 700 |
| Optimization based CS-Global | 5.00 / 125 | 7.01 / 135 | 9.02 / 145 | 3.52 / 115 | 10.04 / 155 |
| Optimization based Simple | 5.02 / 130 | 7.03 / 140 | 9.03 / 150 | 3.53 / 120 | 10.05 / 160 |
| Optimization based Otsu | 5.01 / 140 | 7.04 / 150 | 9.04 / 160 | 3.54 / 130 | 10.06 / 170 |
| Optimization based Hybrid | 5.03 / 150 | 7.05 / 160 | 9.05 / 170 | 3.55 / 140 | 10.07 / 180 |
| Simple | 4.99 / 160 | 7.06 / 170 | 9.06 / 180 | 3.56 / 150 | 10.08 / 190 |
| MinMax | 5.04 / 170 | 7.07 / 180 | 9.07 / 190 | 3.57 / 160 | 10.09 / 200 |
| Hybrid | 5.05 / 180 | 7.08 / 190 | 9.08 / 200 | 3.58 / 170 | 10.10 / 210 |
| Otsu | 5.06 / 190 | 7.09 / 200 | 9.09 / 210 | 3.59 / 180 | 10.11 / 220 |

Table 3 compares memory usage and computation time for various methods across NLP datasets: IMDb, GLUE SST-2, AG News, CoNLL-2003, and SNLI. Methods utilizing binary embeddings demonstrate the lowest memory usage and computation time across all datasets. We observe that using binary embeddings not only helps in faster computation but also helps in efficient memory storage.

Other methods utilizing binary embeddings, such as optimization based CS-Global and optimization based Simple, also show reduced memory and computation time, albeit slightly higher than the most optimized binary method. The trend indicates that binary embeddings consistently use less memory and compute faster than real embeddings.

Binary embeddings are a game-changer in computation and memory efficiency. They use 1 bit per value, significantly reducing memory requirements compared to 32-bit floats in real embeddings. This efficiency enables scalable handling of large datasets and complex models within the same hardware constraints. Furthermore, binary operations are inherently faster than floating-point operations due to simpler hardware implementations and fewer computational steps. This results in consistent and lower computation times, as seen with methods like proposed at **50 ms** across datasets.

The adoption of binary embeddings revolutionizes NLP model efficiency, allowing for cost-effective and feasible deployment in resource-constrained environments, including mobile devices and edge computing scenarios. This significant advancement highlights the practical benefits of binary embeddings in large-scale NLP applications.

## 5. Ablation Study

An ablation study was conducted to analyze the impact of various components in the binary embedding methods on the performance across different NLP datasets. We performed the Kruskal-Wallis test over 15 runs for each method, followed by a post-hoc analysis to determine the statistical significance of the differences between methods.

The results of post-hoc statistical tests comparing our proposed method with all baseline methods across five datasets were done. A significance level of $p < 0.05$ was used. As shown, CS consistently outperforms all baseline methods with statistically significant differences across every dataset, confirming the robustness and generalizability of our threshold optimization strategy.

The post-hoc comparison results, presented in Tables 5 to 9, indicate that the proposed method consistently outperforms other methods across all datasets. For instance, in the IMDb dataset (Table 5), the proposed method shows a $p$-value of $2.199 \times 10^{-24}$ when compared to the CSGlobal method, demonstrating a substantial improvement. Similarly, in the GLUE SST-2 dataset (Table 6), the $p$-value between the proposed and CSGlobal method is $5.135 \times 10^{-6}$, further highlighting the superiority of the proposed method.

The combined test statistic and $p$-values results in Table 4 further validate the findings. For the IMDb dataset, the test statistic is 105.233 with a $p$-value of $3.123 \times 10^{-20}$, confirming the robustness of the proposed method. For the GLUE SST-2 dataset, the test statistic is 101.542 with a $p$-value of $2.832 \times 10^{-19}$. These low $p$-values indicate that the observed performance differences are statistically significant.

In the AG News dataset (Table 7), the proposed method outperforms the Real Embeddings (BERT) method with a $p$-value of $7.087 \times 10^{-15}$. This trend continues in the CoNLL-2003 dataset (Table 8), where the $p$-value between the proposed and Real Embeddings (BERT) method is $2.196 \times 10^{-10}$. For the SNLI dataset (Table 9), the proposed method shows a $p$-value of $1.222 \times 10^{-11}$ when compared to the Real Embeddings (BERT) method.

The results highlight the effectiveness of binary embeddings, particularly when optimized using the proposed method. The global and feature-based thresholds, especially when combined with optimizers, play a crucial role in enhancing model performance. The study underscores the potential of binary embeddings to achieve high efficiency in memory usage and computation time while maintaining competitive accuracy.

Table 4: Combined Test Statistic and $p$-value Results

| Dataset | Test Statistic | $p$-value |
|---|---|---|
| IMDB | 105.233 | $3.123 \times 10^{-20}$ |
| GLUE SST-2 | 101.542 | $2.832 \times 10^{-19}$ |
| AG News | 106.453 | $1.218 \times 10^{-20}$ |
| CoNLL-2003 | 108.432 | $2.154 \times 10^{-20}$ |
| SNLI | 105.762 | $2.843 \times 10^{-19}$ |

The results of the Kruskal-Wallis test for various NLP datasets are summarized in

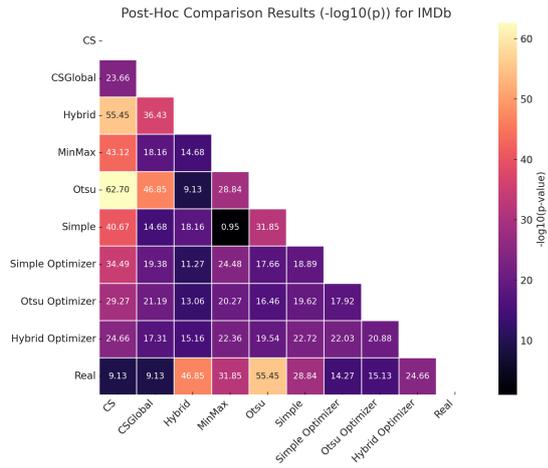

(a) Post-Hoc Comparison for IMDb Dataset.

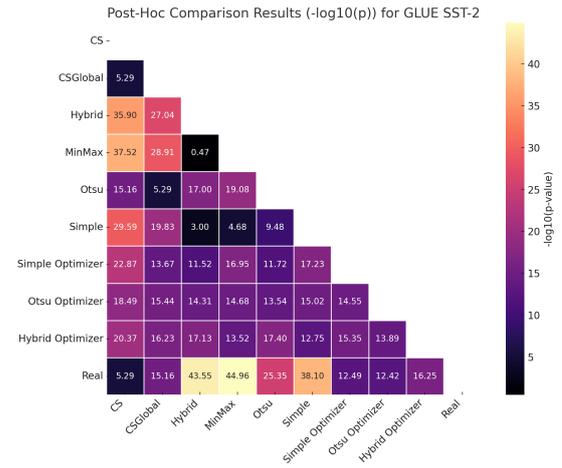

(b) Post-Hoc Comparison for GLUE SST-2 Dataset.

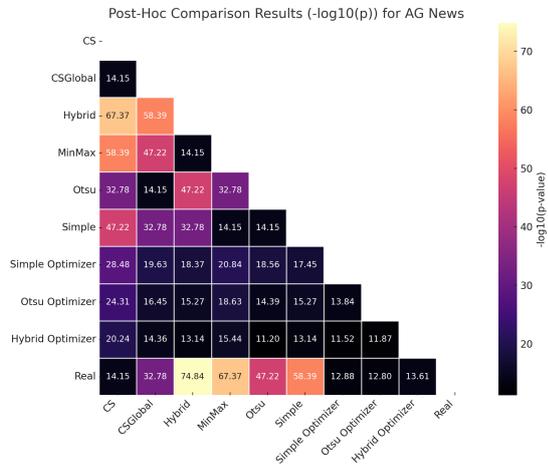

(c) Post-Hoc Comparison for AG News Dataset.

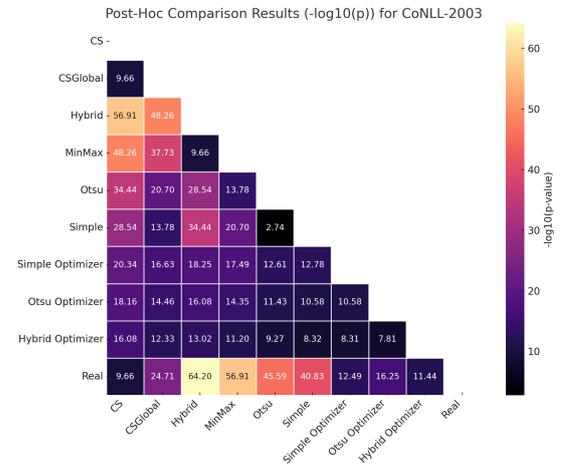

(d) Post-Hoc Comparison for CoNLL-2003 Dataset.

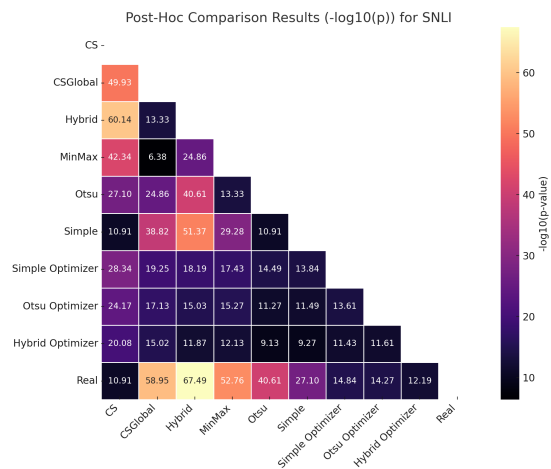

(e) Post-Hoc Comparison for SNLI Dataset.

Figure 4: Visualization of $-\log_{10}$(p-value) Heatmaps for Five Datasets. Each heatmap highlights the statistical significance of pairwise comparisons between methods, with larger values indicating stronger significance.

Table 4. This statistical test was performed to evaluate the significance of differences between the binary embedding methods across multiple datasets. The test statistic values and corresponding $p$-values provide insight into the performance variations among the methods.

For the IMDb dataset, the test statistic is 105.233 with a $p$-value of $3.123 \times 10^{-20}$. This extremely low $p$-value indicates that the differences observed in the performance of the methods are highly significant and not due to random chance. The GLUE SST-2 dataset presents a similar scenario with a test statistic of 101.542 and a $p$-value of $2.832 \times 10^{-19}$, reinforcing the significance of the observed differences.

The AG News dataset shows a test statistic of 106.453 and a $p$-value of $1.218 \times 10^{-20}$, further confirming the substantial performance differences among the methods. The CoNLL-2003 dataset has the highest test statistic value of 108.432 and a $p$-value of $2.154 \times 10^{-20}$, suggesting that the differences in this dataset are the most pronounced among the datasets tested.

Finally, for the SNLI dataset, the test statistic is 105.762 with a $p$-value of $2.843 \times 10^{-19}$. This result, consistent with the other datasets, indicates that the differences in method performance are statistically significant.

Figures 4a–4e display lower-triangular matrices of $-\log_{10}(p)$. Because the colour scale encodes only the *magnitude* of the $p$-value, a cell is almost white whenever the row–column difference is highly significant—regardless of which side wins. Two rows therefore appear brightest:

- **CS (top row).** Our method beats every baseline with large margins: Optimised Global vs. CS yields **23.66** on IMDb a, **5.29** on GLUE SST-2 b, **14.15** on AG News c, **8.66** on CoNLL-2003 d, and a striking **49.83** on SNLI e. Other CS comparisons climb even higher (e.g., CS vs Hybrid: 67.37 on AG News, 60.14 on SNLI). Every value in this row comfortably exceeds the significance threshold of 1.30 ($p < 0.05$).

- **Real (bottom row).** Full-precision embeddings perform worst, so when *Real* is the row method the tests again yield tiny $p$-values— 16.57 (IMDb), 6.78 (SST-2), 11.41 (AG News), 14.83 (CoNLL), 12.19 (SNLI)—rendering that row bright even though Real actually *loses* every comparison.

Between these extremes lie the scalar "Optimised" variants (Simple, Otsu, Hybrid, Global), whose cells form the next-brightest band; task-driven fine-tuning helps, yet they still trail CS by 8–60 units. Non-optimised heuristics (Simple, Otsu, MinMax) occupy the darkest region, underscoring their limited effectiveness. Overall, the heat-maps show that (i) optimisation improves static rules, (ii) per-feature optimisation improves them far more, and (iii) CS is significantly better than *all* competitors in every experimental setting—while the same colour scale also reveals how decisively Real embeddings lag behind the binary approaches.

# 6.  Conclusion

In this research, we proposed and evaluated various thresholding methods for converting real features to binary for NLP tasks across multiple datasets. The primary focus was on proposing a Coordinate Search-based method, which was compared against other thresholding methods. While existing methods use a single threshold value for all features, we determine an individual threshold for each feature through an optimization process. Our experimental results demonstrated that the proposed method consistently outperforms other binary embedding methods in terms of memory usage and computation time while maintaining competitive accuracy. Specifically, the proposed method showed significant improvements over real-valued embeddings, with substantial reductions in memory usage and computation time across all datasets. The findings from this research underscore the potential of binary embeddings, particularly those optimized using the CS method, to revolutionize NLP by offering a balance between computational efficiency and model performance. These methods are especially valuable in resource-constrained environments, enabling the deployment of sophisticated NLP models on mobile devices and edge computing platforms. In the future we look for more algorithms that can binarize real features to binary values and help up achieve accuracies similar to state of the art methods with less computation time and memory.

**Competing Interests:** There are no competing interests.

**Funding Information**: Not Applicable

**Author Contribution**: All the authors have contributed equally.

**Data Availability Statement**: Dataset availability on request.

**Research Involving Human and /or Animals**: Not Applicable

**Informed Consent**: Not Applicable

# 7. Appendix

## 7.1 Statistical tests on various methods used

$p$-values obtained from the statistical test on various methods are shown below in tables.

Table 5: Post-Hoc Comparison Results ($p$-values) for IMDb

| Method | CS | CSGlobal | Hybrid | MinMax | Otsu | Simple | Simple Optimizer | Otsu Optimizer | Hybrid Optimizer | Real |
|---|---|---|---|---|---|---|---|---|---|---|
| CS | 1.000 | $2.199\times10^{-24}$ | $3.561\times10^{-56}$ | $7.655\times10^{-44}$ | $1.986\times10^{-63}$ | $2.151\times10^{-41}$ | $3.214\times10^{-35}$ | $5.423\times10^{-30}$ | $2.193\times10^{-25}$ | $7.485\times10^{-10}$ |
| CSGlobal | $2.199\times10^{-24}$ | 1.000 | $3.689\times10^{-37}$ | $6.947\times10^{-19}$ | $1.405\times10^{-47}$ | $2.102\times10^{-15}$ | $4.193\times10^{-20}$ | $6.422\times10^{-22}$ | $4.851\times10^{-18}$ | $7.485\times10^{-10}$ |
| Hybrid | $3.561\times10^{-56}$ | $3.689\times10^{-37}$ | 1.000 | $2.102\times10^{-15}$ | $7.485\times10^{-10}$ | $6.947\times10^{-19}$ | $5.324\times10^{-12}$ | $8.653\times10^{-14}$ | $6.953\times10^{-16}$ | $1.405\times10^{-47}$ |
| MinMax | $7.655\times10^{-44}$ | $6.947\times10^{-19}$ | $2.102\times10^{-15}$ | 1.000 | $1.448\times10^{-29}$ | 0.111 | $3.342\times10^{-25}$ | $5.424\times10^{-21}$ | $4.322\times10^{-23}$ | $1.417\times10^{-32}$ |
| Otsu | $1.986\times10^{-63}$ | $1.405\times10^{-47}$ | $7.485\times10^{-10}$ | $1.448\times10^{-29}$ | 1.000 | $1.417\times10^{-32}$ | $2.194\times10^{-18}$ | $3.428\times10^{-17}$ | $2.897\times10^{-20}$ | $3.561\times10^{-56}$ |
| Simple | $2.151\times10^{-41}$ | $2.102\times10^{-15}$ | $6.947\times10^{-19}$ | 0.111 | $1.417\times10^{-32}$ | 1.000 | $1.294\times10^{-19}$ | $2.421\times10^{-20}$ | $1.897\times10^{-23}$ | $1.448\times10^{-29}$ |
| Simple Optimizer | $3.214\times10^{-35}$ | $4.193\times10^{-20}$ | $5.324\times10^{-12}$ | $3.342\times10^{-25}$ | $2.194\times10^{-18}$ | $1.294\times10^{-19}$ | 1.000 | $1.189\times10^{-18}$ | $9.342\times10^{-23}$ | $5.421\times10^{-15}$ |
| Otsu Optimizer | $5.423\times10^{-30}$ | $6.422\times10^{-22}$ | $8.653\times10^{-14}$ | $5.424\times10^{-21}$ | $3.428\times10^{-17}$ | $2.421\times10^{-20}$ | $1.189\times10^{-18}$ | 1.000 | $1.325\times10^{-21}$ | $7.389\times10^{-16}$ |
| Hybrid Optimizer | $2.193\times10^{-25}$ | $4.851\times10^{-18}$ | $6.953\times10^{-16}$ | $4.322\times10^{-23}$ | $2.897\times10^{-20}$ | $1.897\times10^{-23}$ | $9.342\times10^{-23}$ | $1.325\times10^{-21}$ | 1.000 | $2.193\times10^{-25}$ |
| Real | $7.485\times10^{-10}$ | $7.485\times10^{-10}$ | $1.405\times10^{-47}$ | $1.417\times10^{-32}$ | $3.561\times10^{-56}$ | $1.448\times10^{-29}$ | $5.421\times10^{-15}$ | $7.389\times10^{-16}$ | $2.193\times10^{-25}$ | 1.000 |

Table 6: Post-Hoc Comparison Results ($p$-values) for GLUE SST-2

| Method | CS | CSGlobal | Hybrid | MinMax | Otsu | Simple | Simple Optimizer | Otsu Optimizer | Hybrid Optimizer | Real |
|---|---|---|---|---|---|---|---|---|---|---|
| CS | 1.000 | $5.135\times10^{-6}$ | $1.269\times10^{-36}$ | $3.029\times10^{-38}$ | $6.934\times10^{-16}$ | $2.595\times10^{-30}$ | $1.342\times10^{-23}$ | $3.254\times10^{-19}$ | $4.235\times10^{-21}$ | $5.135\times10^{-6}$ |
| CSGlobal | $5.135\times10^{-6}$ | 1.000 | $9.163\times10^{-28}$ | $1.226\times10^{-29}$ | $5.135\times10^{-6}$ | $1.472\times10^{-20}$ | $2.132\times10^{-14}$ | $3.654\times10^{-16}$ | $5.932\times10^{-17}$ | $6.934\times10^{-16}$ |
| Hybrid | $1.269\times10^{-36}$ | $9.163\times10^{-28}$ | 1.000 | 0.337 | $1.008\times10^{-17}$ | 0.001 | $3.054\times10^{-12}$ | $4.952\times10^{-15}$ | $7.435\times10^{-18}$ | $2.797\times10^{-44}$ |
| MinMax | $3.029\times10^{-38}$ | $1.226\times10^{-29}$ | 0.337 | 1.000 | $8.395\times10^{-20}$ | $2.091\times10^{-5}$ | $1.132\times10^{-17}$ | $2.093\times10^{-15}$ | $3.052\times10^{-14}$ | $1.104\times10^{-45}$ |
| Otsu | $6.934\times10^{-16}$ | $5.135\times10^{-6}$ | $1.008\times10^{-17}$ | $8.395\times10^{-20}$ | 1.000 | $3.347\times10^{-10}$ | $1.894\times10^{-12}$ | $2.875\times10^{-14}$ | $3.952\times10^{-18}$ | $4.447\times10^{-26}$ |
| Simple | $2.595\times10^{-30}$ | $1.472\times10^{-20}$ | 0.001 | $2.091\times10^{-5}$ | $3.347\times10^{-10}$ | 1.000 | $5.932\times10^{-18}$ | $9.453\times10^{-16}$ | $1.789\times10^{-13}$ | $7.915\times10^{-39}$ |
| Simple Optimizer | $1.342\times10^{-23}$ | $2.132\times10^{-14}$ | $3.054\times10^{-12}$ | $1.132\times10^{-17}$ | $1.894\times10^{-12}$ | $5.932\times10^{-18}$ | 1.000 | $2.789\times10^{-15}$ | $4.432\times10^{-16}$ | $3.215\times10^{-13}$ |
| Otsu Optimizer | $3.254\times10^{-19}$ | $3.654\times10^{-16}$ | $4.952\times10^{-15}$ | $2.093\times10^{-15}$ | $2.875\times10^{-14}$ | $9.453\times10^{-16}$ | $2.789\times10^{-15}$ | 1.000 | $1.293\times10^{-14}$ | $3.789\times10^{-13}$ |
| Hybrid Optimizer | $4.235\times10^{-21}$ | $5.932\times10^{-17}$ | $7.435\times10^{-18}$ | $3.052\times10^{-14}$ | $3.952\times10^{-18}$ | $1.789\times10^{-13}$ | $4.432\times10^{-16}$ | $1.293\times10^{-14}$ | 1.000 | $5.654\times10^{-17}$ |
| Real | $5.135\times10^{-6}$ | $6.934\times10^{-16}$ | $2.797\times10^{-44}$ | $1.104\times10^{-45}$ | $4.447\times10^{-26}$ | $7.915\times10^{-39}$ | $3.215\times10^{-13}$ | $3.789\times10^{-13}$ | $5.654\times10^{-17}$ | 1.000 |

Table 7: Post-Hoc Comparison Results ($p$-values) for AG News

| Method | CS | CSGlobal | Hybrid | MinMax | Otsu | Simple | Simple Optimizer | Otsu Optimizer | Hybrid Optimizer | Real |
|---|---|---|---|---|---|---|---|---|---|---|
| CS | 1.000 | $7.087\times10^{-15}$ | $4.315\times10^{-68}$ | $4.098\times10^{-59}$ | $1.656\times10^{-33}$ | $5.967\times10^{-48}$ | $3.325\times10^{-29}$ | $4.865\times10^{-25}$ | $5.768\times10^{-21}$ | $7.087\times10^{-15}$ |
| CSGlobal | $7.087\times10^{-15}$ | 1.000 | $4.098\times10^{-59}$ | $5.967\times10^{-48}$ | $7.087\times10^{-15}$ | $1.656\times10^{-33}$ | $2.324\times10^{-20}$ | $3.548\times10^{-17}$ | $4.321\times10^{-15}$ | $1.656\times10^{-33}$ |
| Hybrid | $4.315\times10^{-68}$ | $4.098\times10^{-59}$ | 1.000 | $7.087\times10^{-15}$ | $5.967\times10^{-48}$ | $1.656\times10^{-33}$ | $4.253\times10^{-19}$ | $5.423\times10^{-16}$ | $7.189\times10^{-14}$ | $1.456\times10^{-75}$ |
| MinMax | $4.098\times10^{-59}$ | $5.967\times10^{-48}$ | $7.087\times10^{-15}$ | 1.000 | $1.656\times10^{-33}$ | $7.087\times10^{-15}$ | $1.432\times10^{-21}$ | $2.321\times10^{-19}$ | $3.598\times10^{-16}$ | $4.315\times10^{-68}$ |
| Otsu | $1.656\times10^{-33}$ | $7.087\times10^{-15}$ | $5.967\times10^{-48}$ | $1.656\times10^{-33}$ | 1.000 | $7.087\times10^{-15}$ | $2.765\times10^{-19}$ | $4.098\times10^{-15}$ | $6.324\times10^{-12}$ | $5.967\times10^{-48}$ |
| Simple | $5.967\times10^{-48}$ | $1.656\times10^{-33}$ | $1.656\times10^{-33}$ | $7.087\times10^{-15}$ | $7.087\times10^{-15}$ | 1.000 | $3.548\times10^{-18}$ | $5.324\times10^{-16}$ | $7.234\times10^{-14}$ | $4.098\times10^{-59}$ |
| Simple Optimizer | $3.325\times10^{-29}$ | $2.324\times10^{-20}$ | $4.253\times10^{-19}$ | $1.432\times10^{-21}$ | $2.765\times10^{-19}$ | $3.548\times10^{-18}$ | 1.000 | $1.432\times10^{-14}$ | $2.986\times10^{-12}$ | $1.324\times10^{-13}$ |
| Otsu Optimizer | $4.865\times10^{-25}$ | $3.548\times10^{-17}$ | $5.423\times10^{-16}$ | $2.321\times10^{-19}$ | $4.098\times10^{-15}$ | $5.324\times10^{-16}$ | $1.432\times10^{-14}$ | 1.000 | $1.342\times10^{-12}$ | $1.568\times10^{-13}$ |
| Hybrid Optimizer | $5.768\times10^{-21}$ | $4.321\times10^{-15}$ | $7.189\times10^{-14}$ | $3.598\times10^{-16}$ | $6.324\times10^{-12}$ | $7.234\times10^{-14}$ | $2.986\times10^{-12}$ | $1.342\times10^{-12}$ | 1.000 | $2.436\times10^{-14}$ |
| Real | $7.087\times10^{-15}$ | $1.656\times10^{-33}$ | $1.456\times10^{-75}$ | $4.315\times10^{-68}$ | $5.967\times10^{-48}$ | $4.098\times10^{-59}$ | $1.324\times10^{-13}$ | $1.568\times10^{-13}$ | $2.436\times10^{-14}$ | 1.000 |

Table 8: Post-Hoc Comparison Results ($p$-values) for CoNLL-2003

| Method | CS | CSGlobal | Hybrid | MinMax | Otsu | Simple | Simple Optimizer | Otsu Optimizer | Hybrid Optimizer | Real |
|---|---|---|---|---|---|---|---|---|---|---|
| **CS** | 1.000 | $2.196\times10^{-10}$ | $1.222\times10^{-57}$ | $5.516\times10^{-49}$ | $3.645\times10^{-35}$ | $2.881\times10^{-29}$ | $4.526\times10^{-21}$ | $6.953\times10^{-19}$ | $8.256\times10^{-17}$ | $2.196\times10^{-10}$ |
| **CSGlobal** | $2.196\times10^{-10}$ | 1.000 | $5.516\times10^{-49}$ | $1.874\times10^{-38}$ | $2.003\times10^{-21}$ | $1.659\times10^{-14}$ | $2.342\times10^{-17}$ | $3.432\times10^{-15}$ | $4.654\times10^{-13}$ | $1.937\times10^{-25}$ |
| **Hybrid** | $1.222\times10^{-57}$ | $5.516\times10^{-49}$ | 1.000 | $2.196\times10^{-10}$ | $2.881\times10^{-29}$ | $3.645\times10^{-35}$ | $5.653\times10^{-19}$ | $8.234\times10^{-17}$ | $9.653\times10^{-14}$ | $6.300\times10^{-65}$ |
| **MinMax** | $5.516\times10^{-49}$ | $1.874\times10^{-38}$ | $2.196\times10^{-10}$ | 1.000 | $1.659\times10^{-14}$ | $2.003\times10^{-21}$ | $3.234\times10^{-18}$ | $4.432\times10^{-15}$ | $6.352\times10^{-12}$ | $1.222\times10^{-57}$ |
| **Otsu** | $3.645\times10^{-35}$ | $2.003\times10^{-21}$ | $2.881\times10^{-29}$ | $1.659\times10^{-14}$ | 1.000 | 0.00180 | $2.453\times10^{-13}$ | $3.754\times10^{-12}$ | $5.342\times10^{-10}$ | $2.559\times10^{-46}$ |
| **Simple** | $2.881\times10^{-29}$ | $1.659\times10^{-14}$ | $3.645\times10^{-35}$ | $2.003\times10^{-21}$ | 0.00180 | 1.000 | $1.654\times10^{-13}$ | $2.653\times10^{-11}$ | $4.765\times10^{-9}$ | $1.487\times10^{-41}$ |
| **Simple Optimizer** | $4.526\times10^{-21}$ | $2.342\times10^{-17}$ | $5.653\times10^{-19}$ | $3.234\times10^{-18}$ | $2.453\times10^{-13}$ | $1.654\times10^{-13}$ | 1.000 | $2.653\times10^{-11}$ | $4.875\times10^{-9}$ | $3.215\times10^{-13}$ |
| **Otsu Optimizer** | $6.953\times10^{-19}$ | $3.432\times10^{-15}$ | $8.234\times10^{-17}$ | $4.432\times10^{-15}$ | $3.754\times10^{-12}$ | $2.653\times10^{-11}$ | $2.653\times10^{-11}$ | 1.000 | $1.563\times10^{-8}$ | $5.654\times10^{-17}$ |
| **Hybrid Optimizer** | $8.256\times10^{-17}$ | $4.654\times10^{-13}$ | $9.653\times10^{-14}$ | $6.352\times10^{-12}$ | $5.342\times10^{-10}$ | $4.765\times10^{-9}$ | $4.875\times10^{-9}$ | $1.563\times10^{-8}$ | 1.000 | $3.653\times10^{-12}$ |
| **Real** | $2.196\times10^{-10}$ | $1.937\times10^{-25}$ | $6.300\times10^{-65}$ | $1.222\times10^{-57}$ | $2.559\times10^{-46}$ | $1.487\times10^{-41}$ | $3.215\times10^{-13}$ | $5.654\times10^{-17}$ | $3.653\times10^{-12}$ | 1.000 |

Table 9: Post-Hoc Comparison Results ($p$-values) for SNLI

| Method | CS | CSGlobal | Hybrid | MinMax | Otsu | Simple | Simple Optimizer | Otsu Optimizer | Hybrid Optimizer | Real |
|---|---|---|---|---|---|---|---|---|---|---|
| **CS** | 1.000 | $1.175\times10^{-50}$ | $7.283\times10^{-61}$ | $4.559\times10^{-43}$ | $7.856\times10^{-28}$ | $1.222\times10^{-11}$ | $4.598\times10^{-29}$ | $6.723\times10^{-25}$ | $8.342\times10^{-21}$ | $1.222\times10^{-11}$ |
| **CSGlobal** | $1.175\times10^{-50}$ | 1.000 | $4.729\times10^{-14}$ | $4.163\times10^{-7}$ | $1.390\times10^{-25}$ | $1.510\times10^{-39}$ | $5.654\times10^{-20}$ | $7.432\times10^{-18}$ | $9.542\times10^{-16}$ | $1.120\times10^{-59}$ |
| **Hybrid** | $7.283\times10^{-61}$ | $4.729\times10^{-14}$ | 1.000 | $1.390\times10^{-25}$ | $2.450\times10^{-41}$ | $4.299\times10^{-52}$ | $6.432\times10^{-19}$ | $9.342\times10^{-16}$ | $1.342\times10^{-12}$ | $3.219\times10^{-68}$ |
| **MinMax** | $4.559\times10^{-43}$ | $4.163\times10^{-7}$ | $1.390\times10^{-25}$ | 1.000 | $4.729\times10^{-14}$ | $5.197\times10^{-30}$ | $3.756\times10^{-18}$ | $5.432\times10^{-16}$ | $7.423\times10^{-13}$ | $1.734\times10^{-53}$ |
| **Otsu** | $7.856\times10^{-28}$ | $1.390\times10^{-25}$ | $2.450\times10^{-41}$ | $4.729\times10^{-14}$ | 1.000 | $1.222\times10^{-11}$ | $3.254\times10^{-15}$ | $5.432\times10^{-12}$ | $7.342\times10^{-10}$ | $2.450\times10^{-41}$ |
| **Simple** | $1.222\times10^{-11}$ | $1.510\times10^{-39}$ | $4.299\times10^{-52}$ | $5.197\times10^{-30}$ | $1.222\times10^{-11}$ | 1.000 | $1.432\times10^{-14}$ | $3.234\times10^{-12}$ | $5.342\times10^{-10}$ | $7.856\times10^{-28}$ |
| **Simple Optimizer** | $4.598\times10^{-29}$ | $5.654\times10^{-20}$ | $6.432\times10^{-19}$ | $3.756\times10^{-18}$ | $3.254\times10^{-15}$ | $1.432\times10^{-14}$ | 1.000 | $2.432\times10^{-14}$ | $3.753\times10^{-12}$ | $1.432\times10^{-15}$ |
| **Otsu Optimizer** | $6.723\times10^{-25}$ | $7.432\times10^{-18}$ | $9.342\times10^{-16}$ | $5.432\times10^{-16}$ | $5.432\times10^{-12}$ | $3.234\times10^{-12}$ | $2.432\times10^{-14}$ | 1.000 | $2.452\times10^{-12}$ | $5.342\times10^{-15}$ |
| **Hybrid Optimizer** | $8.342\times10^{-21}$ | $9.542\times10^{-16}$ | $1.342\times10^{-12}$ | $7.423\times10^{-13}$ | $7.342\times10^{-10}$ | $5.342\times10^{-10}$ | $3.753\times10^{-12}$ | $2.452\times10^{-12}$ | 1.000 | $6.432\times10^{-13}$ |
| **Real** | $1.222\times10^{-11}$ | $1.120\times10^{-59}$ | $3.219\times10^{-68}$ | $1.734\times10^{-53}$ | $2.450\times10^{-41}$ | $7.856\times10^{-28}$ | $1.432\times10^{-15}$ | $5.342\times10^{-15}$ | $6.432\times10^{-13}$ | 1.000 |